\documentclass{article}
\usepackage{ijcai20}

\usepackage{times}
\usepackage{soul}
\usepackage{url}
\usepackage[hidelinks]{hyperref}
\usepackage[utf8]{inputenc}
\usepackage[small]{caption}
\usepackage{graphicx}
\usepackage{amsmath}
\usepackage{amsthm}
\usepackage{booktabs}
\usepackage{algorithm}
\usepackage{algorithmic}
\usepackage{epsfig}
\usepackage{graphicx}
\usepackage{amsmath}
\usepackage{amssymb}
\usepackage{bm}
\usepackage{caption}
\usepackage{graphicx, subfig}
\usepackage{verbatim}

\usepackage{graphicx,times}  
\usepackage{subfig} 
\usepackage{graphicx} 
\usepackage{color}

\usepackage{amsthm,amsmath,amssymb}
\usepackage{mathrsfs}

\title{Network Cooperation with Progressive Disambiguation \\ for Partial Label Learning}

\author{
\author{
	Yao Yao\textsuperscript{\rm 1}, 
	Chen Gong\textsuperscript{\rm 1}, 
	Jiehui Deng\textsuperscript{\rm 1}, 
	Jian Yang\textsuperscript{\rm 1,2}\\
	\textsuperscript{\rm 1}Nanjing University of Science and Technology, China\\ 
	\textsuperscript{\rm 2}Jiangsu Key Lab of Image and Video Understanding for Social Security\\
	\{yaoyao, chen.gong, jhdeng, csjyang\}@njust.edu.cn
}
}

\begin{document}

\maketitle

\begin{abstract}
Partial Label Learning (PLL) aims to train a classifier when each training instance is associated with a set of candidate labels, among which only one is correct but is not accessible during the training phase. The common strategy dealing with such ambiguous labeling information is to disambiguate the candidate label sets. Nonetheless, existing methods ignore the disambiguation difficulty of instances and adopt the single-trend training mechanism. The former would lead to the vulnerability of models to the false positive labels and the latter may arouse error accumulation problem. To remedy these two drawbacks, this paper proposes a novel approach termed ``Network Cooperation with Progressive Disambiguation'' (NCPD) for PLL. Specifically, we devise a progressive disambiguation strategy of which the disambiguation operations are performed on simple instances firstly and then gradually on more complicated ones. Therefore, the negative impacts brought by the false positive labels of complicated instances can be effectively mitigated as the disambiguation ability of the model has been strengthened via learning from the simple instances. Moreover, by employing artificial neural networks as the backbone, we utilize a network cooperation mechanism which trains two networks collaboratively by letting them interact with each other. As two networks have different disambiguation ability, such interaction is beneficial for both networks to reduce their respective disambiguation errors, and thus is much better than the existing algorithms with single-trend training process. Extensive experimental results on various benchmark and practical datasets demonstrate the superiority of our NCPD to other state-of-the-art PLL methods.

\end{abstract}

\section{Introduction}
\label{introduction}
Partial Label Learning (PLL), which is also known as \emph{superset label learning}~\cite{liu2012conditional,gong2018regular} and \emph{ambiguous label learning}~\cite{hullermeier2006,chen2014ambiguously}, is one of the emerging research fields in weakly-supervised learning. PLL learns from ambiguous labeling information where each training instance is associated with multiple candidate labels and only one of them is valid. Due to the prevalence of ambiguous labeling in real-world scenarios, PLL has many practical applications such as image annotation~\cite{cour2009learning,chen2018learning}, ecoinformatics~\cite{liu2012conditional}, web mining~\cite{luo2010learning}, etc.

Formally, let $\mathcal{X}\in\mathbb{R}^{d}$ denote the \emph{d}-dimensional input space and $\mathcal{Y}= \{1, 2, \cdots, c \}$ denote the label space with \emph{c} class labels. The task of PLL is to induce a classifier $f:\mathcal{X}\rightarrow\mathcal{Y}$ from the partial label training set $\mathcal{D}=\{(\mathbf{x}_i,S_i)| 1\le i \le N\}$, where $\mathbf{x}_i \in \mathcal{X}$ is a $d$-dimensional feature vector and $S_i \subseteq \mathcal{Y}$ is the corresponding candidate label set of $\mathbf{x}_i$. Particularly, the basic assumption under PLL framework is that the latent groundtruth label ${\rm{y}}_i$ of $\mathbf{x}_i$ lies in $S_i$, \emph{i.e.}, ${\rm{y}}_i \in S_i$, whereas it is not directly accessible during the training phase.

To learn from such partially labeled instances with ambiguously supervised information, the common strategy is to disambiguate the set of candidate labels of each training instance, namely to detect the unique correct label among multiple candidate labels. There are mainly two classes of methods for such disambiguation operation, namely average-based methods and identification-based methods. Average-based methods treat all candidate labels equally by assuming that they contribute equally to the trained classifier and the prediction is made by averaging their model outputs~\cite{hullermeier2006,zhang2015solving}. These methods share a common deficiency that the effectiveness of the model is greatly affected by the false positive labels in the candidate label sets, which leads to the suppression of groundtruth label by these false positive labels. Identification-based methods address this shortcoming via considering groundtruth label as a latent variable and gradually identifying it by iterative procedures such as Expectation Maximization (EM)~\cite{jin2003learning,nguyen2008classification,yu2017maximum}. One potential drawback of identification-based methods is that rather than recovering the latent groundtruth labels, the identified labels might turn out to be false positive and they can hardly be rectified in the subsequent iterations.

In a word, existing methods are vulnerable to false positive labels in the candidate label sets. There are two critical reasons that account for this. Firstly, existing approaches scarcely take the disambiguation difficulty of instances into account, and the disambiguation operations are performed on every training instance all at once. In this case, when the instance is complicated and difficult to classify, their models are likely to mistakenly regard the false positive label as the latent groundtruth label, which will mislead the training process and ultimately impair the disambiguation ability of the models. Secondly, the training process of existing methods are all single-trend, which indicates that the data disambiguated at the current step will be directly transferred back to the model itself in the following steps. Under this circumstance, once the identified labels turn out to be false positive, they would be difficult to correct in the succeeding iterations and thereby raising the error accumulation problem, which will severely degrade their performances.

To address these two shortcomings, this paper proposes a novel approach which employs a progressive disambiguation strategy combined with a network cooperation mechanism for PLL, which is termed ``\textbf{N}etwork \textbf{C}ooperation with \textbf{P}rogressive \textbf{D}isambiguation" (``NCPD" for short). Specifically, to address the problem of ignoring the disambiguation difficulty of instances, we devise a progressive disambiguation strategy which disambiguates simple instances firstly and then gradually disambiguates more complicated ones. Through learning from the simple instances, the disambiguation ability of the model can be improved steadily. With the proceeding of training process, the model is capable of disambiguating the complicated instances precisely. As a consequence, the negative impacts brought by the false positive labels, especially those of complicated instances, can be effectively mitigated. To settle the error accumulation problem caused by the single-trend training mechanism of traditional methods, we employ Artificial Neural Networks (ANNs) as the backbone and utilize a network cooperation mechanism which trains two networks collaboratively by letting them interact with each other. That is to say, two networks disambiguate the training instances independently in the forward propagation phase and then back propagate the data disambiguated by its peer network. As two networks have different ability and can disambiguate training instances at different levels, such interaction is beneficial for both networks to learn from each other and thus their respective disambiguation errors can be reduced. As a result, the error accumulation problem can be significantly alleviated, and that is why we adopt such network cooperation mechanism rather than the existing single-trend training process. Intensive experiments on multiple datasets substantiate the superiority of our proposed NCPD approach to the state-of-the-art methodologies.

The rest of this paper is organized as follows. We review the related works in Section~\ref{Related Work}, and introduce the proposed NCPD approach in Section~\ref{method}. Section~\ref{experiments} reports the experimental results, followed by the conclusion in Section~\ref{conclusion}.

\section{Related Work}
\label{Related Work}
Existing algorithms dealing with partially labeled instances can be roughly grouped into the following two classes, \emph{i.e.}, average-based methods and identification-based methods. 

The average-based methods treat all candidate labels equally and the prediction is made by averaging their model outputs. For example, the work~\cite{hullermeier2006} straightforwardly generalizes the $k$-nearest neighbor classifier to resolve the PLL problem by predicting the label of a test instance $\mathbf{x}$ via the voting strategy among the candidate labels of its neighbors. That is to say, $f(\mathbf{x})={\rm{argmax}}{_{y\in\mathcal{Y}}}\sum\nolimits_{i\in\mathcal{N}_{(\mathbf{x})}}\mathbb{I}(y\in S_i)$, where $\mathcal{N}(\mathbf{x})$ denotes the neighbors of the test instance $\mathbf{x}$ and $\mathbb{I}(\cdot)$ is the indicator function. Zhang \emph{et al.}~\cite{zhang2015solving} also propose a model of which the predictions of unseen instances are made by the weighted averaging over the candidate labels of their neighbors. Cour \emph{et al.}~\cite{cour2009learning} propose a convex learning method and decide the groundtruth label by averaging the outputs from all candidate labels, \emph{i.e.}, $\frac{1}{|S_i|}\sum\nolimits_{y\in S_i}F(\mathbf{x},\Theta,y)$ with $\Theta$ being the model parameters. Average-based methods are intuitive and are easy to implement. However, these methods share a critical shortcoming that the outputs from false positive labels may overwhelm the groundtruth labels' outputs, which will severely degrade their performances.

The identification-based methods regard the unique groundtruth label as a latent variable and identify it as ${\rm{argmax}}_{y\in S_i}F(\mathbf{x}_i,\Theta,y)$. Maximum likelihood criterion and maximum margin criterion are the two most widely-used learning strategies to identify groundtruth labels. Based on EM procedure, these methods~\cite{jin2003learning,liu2012conditional} train their models by optimizing the maximum likelihood function $\sum_{i=1}^{n}{\log}(\sum_{y\in S_i}F(\mathbf{x}_i,\Theta,y))$. The work~\cite{nguyen2008classification} maximizes the margin between outputs from candidate labels and that from non-candidate labels to refine groundtruth labels, and the corresponding objective function is $\sum_{i=1}^{n}({\rm{max}}_{y\in S_i}F(\mathbf{x}_i,\Theta,y)-{\rm{max}}_{y\notin S_i}F(\mathbf{x}_i,\Theta,y))$. Nonetheless, the above margin ignores the predictive difference between the latent groundtruth label and other candidate labels. To address this problem, Yu \emph{et al.}~\cite{yu2017maximum} directly maximize the margin between the groundtruth label and other labels, \emph{i.e.}, $\sum_{i=1}^{n}(F(\mathbf{x}_i,\Theta,{{\rm{y}}_i})-{\rm{max}}_{y\neq {{\rm{y}}_i}}F(\mathbf{x}_i,\Theta,y))$ where ${\rm{y}}_i$ denotes the groundtruth label of $\mathbf{x}_i$. Differently, Feng \emph{et al.}~\cite{feng2019partial} balance the minimum approximation loss and the maximum infinity norm of the outputs to differentiate the unique groundtruth label from false positive labels. One potential shortcoming of identification-based methods is that the identified label in the current iteration may turn out to be false positive and they can hardly be rectified in the subsequent iterations.


Although the aforementioned methods have achieved good performances to some degree, they still suffer from two severe drawbacks, \emph{i.e.}, ignoring the disambiguation difficulty of instances and adopting the unreliable single-trend training process, and both of them will degrade their performances as mentioned in the introduction. Therefore, this paper presents a novel algorithm termed NCPD which will be introduced in the next section.

\begin{figure*}[!t]
	\setlength{\belowcaptionskip}{-12 pt}
	\setlength{\abovecaptionskip}{4 pt}
	\centering
	\includegraphics[width=1.45\columnwidth]{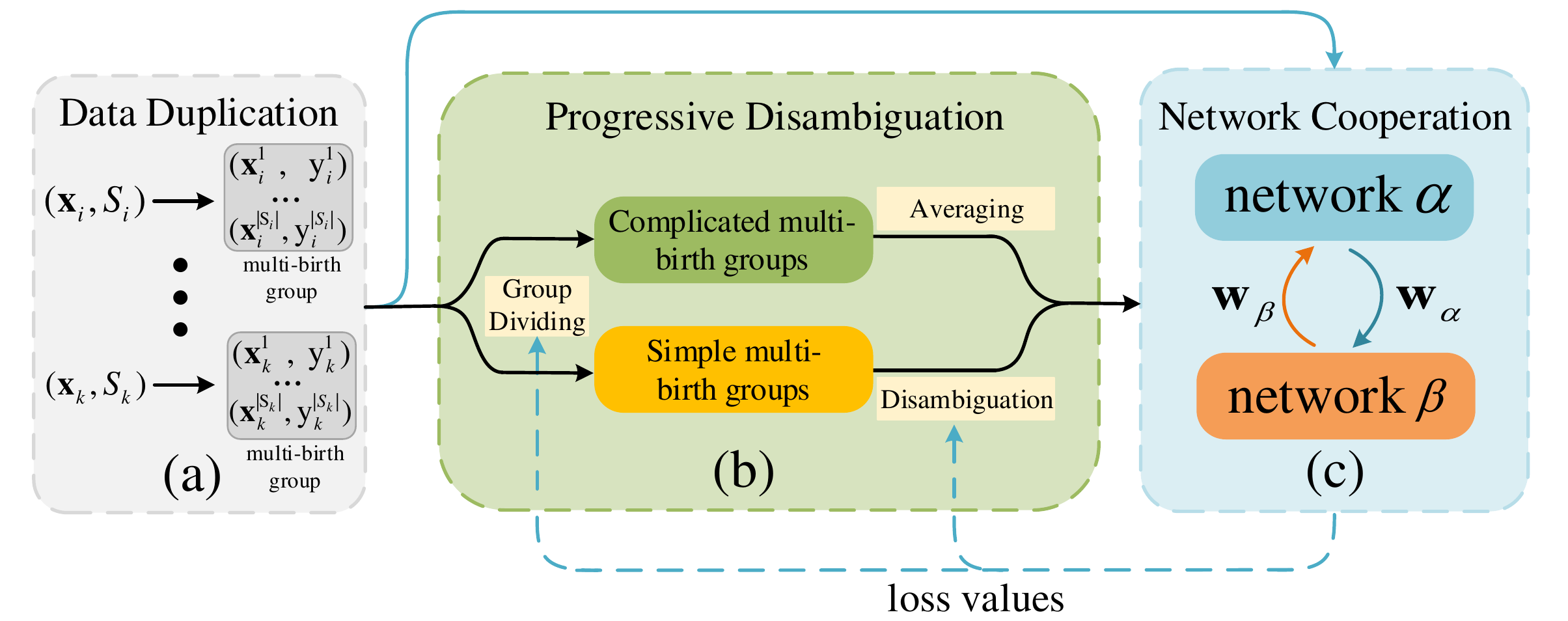}
	\caption{The framework of our method. (a) indicates the data duplication scheme which transforms each partially labeled instance into a multi-birth group. After that, we feed the transformed data into the networks and thus their corresponding loss values can be obtained (the blue line). (b) presents the process of dividing multi-birth groups into two levels of difficulty and then calculating the confidence scores of instances among them according to the incurred loss values. (c) denotes the network cooperation mechanism where two networks interact with each other via exchanging their respective confidence scores of instances (\emph{i.e.}, $\mathbf{w}_\alpha$ and $\mathbf{w}_\beta$) for back propagation. }
	\label{pipline}
\end{figure*}

\section{The Proposed NCPD Approach}
\label{method}
In this section, we introduce the NCPD approach of which the architecture is illustrated in Figure~\ref{pipline}. We firstly employ a data duplication scheme which transforms each partially labeled instance into a multi-birth group\footnote{The notion of ``multi-birth group'' will be detailed later in Section~\ref{3.1}.} (Figure~\ref{pipline} (a)). Afterwards, by dividing these multi-birth groups into two levels of difficulty (\emph{i.e.}, ``simple'' and ``complicated''), we can calculate the confidence scores of instances among them via averaging or disambiguation (Figure~\ref{pipline} (b)). Finally, two networks collaborate with each other through exchanging the confidence scores of instances generated by them independently to compute their respective back propagated loss (Figure~\ref{pipline} (c)). We will detail these critical steps in the following sections. 

\subsection{Data Duplication}
\label{3.1}
We denote $\mathbf{X} =[\mathbf{x}_1,\dots,\mathbf{x}_N]$ as the training set with each column $\mathbf{x}_i$ ($i=1,2,\dots,N$) representing the feature vector of the $i$-th instance and $N$ denotes the total number of training instances. Besides, we represent the candidate label set of $\mathbf{x}_i$ as $S_i=\{{\rm{y}}_i^1, {\rm{y}}_i^2,\dots,{\rm{y}}_i^{|S_i|} \}$, where $|S_i|$ denotes the cardinality of $S_i$.

To pave the way for subsequent disambiguation operations, we adopt a data duplication scheme on the original partially labeled training dataset. Specifically, for an arbitrary training instance $\mathbf{x}_i$ and its corresponding candidate label set $S_i$, we first duplicate $\mathbf{x}_i$ into $|S_i|$ replicas, \emph{i.e.}, $\mathbf{x}_i^1$, $\mathbf{x}_i^2$, $\dots$, and $\mathbf{x}_i^{|S_i|}$, and each replica is identical to the original feature vector $\mathbf{x}_i$. After that, we decompose the corresponding candidate label set $S_i=\{{\rm{y}}_i^1, {\rm{y}}_i^2,\dots,{\rm{y}}_i^{|S_i|} \}$ and then assign each candidate label ${\rm{y}}_i^j$ $(j=1,2,\dots,|S_i|)$ to a replica $\mathbf{x}_i^j$. Eventually, from an original training instance $\mathbf{x}_i$ and its corresponding candidate label set $S_i$, we can obtain $|S_i|$ newly generated instance-label pairs, \emph{i.e.}, $(\mathbf{x}_i^1,{\rm{y}}_i^1)$, $(\mathbf{x}_i^2,{\rm{y}}_i^2)$, $\dots$, $(\mathbf{x}_i^{|S_i|},{\rm{y}}_i^{|S_i|})$, and we name these pairs which are generated from the original one instance as a ``\emph{multi-birth group}". 

After performing the above-mentioned data duplication operation on all training instances, we have transformed the original partially labeled training dataset into a new training dataset which contains $n=\sum\nolimits_{i}|S_i|$ ($i=1,2,\dots,N$) instances from $N$ multi-birth groups, and meanwhile each instance contains only one label (can be correct or incorrect). It is worth noting that although learning from such transformed dataset is similar to corrupted labels learning~\cite{gong2017learning,Yi_2019_CVPR} at the first glance, it differs from corrupted label learning in that we can definitely know that only one instance is labeled correctly while the labels of other instances are all wrong among each multi-birth group.

As we have obtained the new training dataset, disambiguating the original partially labeled instances is transformed to disambiguating the multi-birth groups, \emph{i.e.}, detecting the unique correctly labeled instance in each multi-birth group. To achieve this target, we take the confidence level of each training instance into consideration. Specifically, we denote $\mathbf{w}={[\mathbf{w}_1^\top,\mathbf{w}_2^\top,\dots,\mathbf{w}_N^\top]}^\top\in\mathbb{R}^{n\times1}$ as the confidence vector of $n$ training instances from $N$ multi-birth groups, where $\mathbf{w}_i={[w_i^1,w_i^2,\dots,w_i^{|S_i|}]}^\top$  indicates the group confidence vector of the $i$-th multi-birth group with the $j$-th element $w_i^j\in [0,1]$ in $\mathbf{w}_i$ representing the learning confidence score of the instance $\mathbf{x}_i^j$. As there is only one instance labeled correctly in each multi-birth group, the instances in the same multi-birth group are naturally in a competitive relationship. Therefore, we assume that each group confidence vector should be normalized, \emph{i.e.}, $\sum_{j=1}^{|S_i|}w_i^j=1,\forall i=1,2,\dots,N$. Distinctly, disambiguating the multi-birth groups is equivalent to refining their corresponding group confidence vectors.

\subsection{Progressive Disambiguation}
\label{3.2}
As stated before, we attempt to disambiguate the simple multi-birth groups at the initial training stages and gradually disambiguate more complicated ones as the training process goes on. That is to say, the group confidence vectors of the simple multi-birth groups ought to be acquired firstly so that the trained model is capable of learning from these disambiguated multi-birth groups. With the proceeding of training process, the disambiguation ability of the model will be improved and thus the group confidence vectors of the complicated multi-birth groups can be obtained precisely.

Intuitively, if a multi-birth group contains an instance which is probably labeled correctly, disambiguating this multi-birth group is relatively easy and thus we consider it as a simple multi-birth group. Existing researches~\cite{zhang2016understanding,arpit2017closer} have shown that ANNs will learn clean and easy patterns firstly, which indicates that the instances with small loss values are likely to be correctly labeled. Based on such observation and meanwhile employing ANNs as the backbone, we propose a progressive disambiguation strategy as follows. 

Specifically, after feeding the mini-batch data $\mathcal{D}^b$ into the network at the $t$-th epoch, we can obtain the cross-entropy loss values of these instances, namely $\bm{\ell}(\Theta,\mathcal{D}^b)$, where $\Theta$ indicates the network parameters. After that, we pick up the instances which are likely to be correctly labeled according to the following two conditions: 1) Their loss values are the first $T(t)$ percentage minimums out of $\bm\ell(\Theta,\mathcal{D}^b)$, where $T(t)$ is a time-dependent parameter determining the maximum amount of the simple multi-birth groups at the $t$-th epoch, and we will introduce it later; and 2) They must be predicted correctly, \emph{i.e.}, the network predictions on them are identical to their labels. After the above screening operation, we can fetch several small-loss instances from the mini-batch $\mathcal{D}^b$ and we regard them as \emph{reliable instances}. It is worth noting that each multi-birth group contains at most one reliable instance because of the constraint from the second condition. Next, we can divide multi-birth groups into two levels of difficulty according to whether they contain a reliable instance, namely simple multi-birth groups and complicated multi-birth groups. Each simple multi-birth group contains one reliable instance which is likely to be correctly labeled, and thus we consider this multi-birth group is relatively easy to disambiguate at the current epoch. Therefore, we disambiguate it by assigning distinguishing confidence scores to the instances among it according to their loss values. If the $i$-th multi-birth group is a simple multi-birth group, its corresponding group confidence vector $\mathbf{w}_i$ can be updated as: 
\begin{equation}
\setlength{\abovedisplayskip} {1pt}
\setlength{\belowdisplayskip} {1pt}
w_i^j= \frac{exp(-\ell_i^j)}{\sum_{k=1}^{|S_i|}exp(-\ell_i^k)},j=1,2,\dots,|S_i|, 
\label{map1}
\end{equation}
where $\ell_i^j$ ($\ell_i^k$) indicates the loss value of the $j$-th ($k$-th) instance in the $i$-th multi-birth group. Eq.~(\ref{map1}) indicates that the instances with small loss values can acquire relatively large confidence scores and meanwhile the normalization constraints of group confidence vectors can be satisfied. As to the complicated multi-birth groups which do not contain any reliable instance, we assign an average confidence vector to them as we cannot figure out the correctly labeled instances among them, namely: 
\begin{equation}
\setlength{\abovedisplayskip} {1pt}
\setlength{\belowdisplayskip} {1pt}
w_i^j= \frac{1}{|S_i|},j=1,2,\dots,|S_i|.
\label{map2}
\end{equation}
As no loss value will be generated before the first epoch, all group confidence vectors are initialized in an average manner according to Eq.~(\ref{map2}).

After we have obtained the group confidence vector of each multi-birth group, we can clearly know that the instances with large confidence scores are likely to be correctly labeled, and thereby the trained network should pay more attention to them. Otherwise, the network ought to avoid learning from these instances. Taking this into account, we assign weights to the loss values of the instances (\emph{i.e.}, $\bm{\ell}(\Theta,\mathcal{D}^b$)) with their respective confidence scores, and the propagated back loss of $\mathcal{D}^b$, \emph{i.e.}, $\mathcal{L}(\Theta,\mathcal{D}^b)$, can be calculated as follows:
\begin{equation}
\setlength{\abovedisplayskip} {2pt}
\setlength{\belowdisplayskip} {2pt}
\mathcal{L}(\Theta,\mathcal{D}^b)={\mathbf{w}^b}^\top\bm{\ell}(\Theta,\mathcal{D}^b),
\label{loss}
\end{equation}
where $\mathbf{w}^b$ is the confidence vector concatenated by the confidence scores of instances in $\mathcal{D}^b$. Finally, by denoting $\eta$ as the learning rate, the network parameters $\Theta$ can be updated as:
\begin{equation}
\setlength{\abovedisplayskip} {2pt}
\setlength{\belowdisplayskip} {2pt}
\Theta:=\Theta-\eta\nabla\mathcal{L}(\Theta,\mathcal{D}^b)
\label{theta}.
\end{equation}

As mentioned previously, $T(t)$ is a time-dependent parameter which implies that at most $T(t)$ percentage of multi-birth groups will be regarded as simple multi-birth groups and disambiguated at the $t$-th epoch, and it will increase from zero to one as the training process proceeds. The concrete formulation of $T(t)$ is as follows:
\begin{equation}
\setlength{\abovedisplayskip} {2pt}
\setlength{\belowdisplayskip} {2pt}
T(t)=
\begin{cases}
	exp(-5{({t}/{t_r}-1)}^2)& \ t\leq t_r\\
	 \qquad \qquad  1& \ t> t_r
\end{cases}
,
\label{R}
\end{equation}
where $t_r$ is a coefficient determining at which epoch $T(t)$ reaches to one, meaning that almost all the multi-birth groups will be disambiguated after that epoch. Eq.~(\ref{R}) reveals that at the initial training phase, only very few yet simple multi-birth groups will be disambiguated as $T(t)$ is relatively small. With the advance of training steps, the network disambiguation ability will be strengthened and it is capable of disambiguating the complicated multi-birth groups, and thereby $T(t)$ ought to increase accordingly.

\subsection{Network Cooperation}
Although the aforementioned progressive disambiguation strategy has taken the disambiguation difficulty of multi-birth groups into consideration, the corresponding training process is still single-trend of which the disambiguated data will be directly transfered back to the model itself, and the accompanied shortcomings have been analyzed before. Inspired by the work~\cite{han2018co} dealing with corrupted label learning problem, we devise a network cooperation mechanism, which trains two networks collaboratively and lets them interact with each other regarding the confidence levels of the instances. 

By denoting the two networks as $\alpha$ (with parameter $\Theta_\alpha$) and $\beta$ (with parameter $\Theta_\beta$) respectively, we can obtain two confidence vectors of $\mathcal{D}^b$ generated by them independently (according to Section~\ref{3.2}), \emph{i.e.}, $\mathbf{w}^b_\alpha$ and $\mathbf{w}^b_\beta$. After that, we exchange the confidence vectors among two networks to calculate their respective back propagated loss, \emph{i.e.}, $\mathcal{L}_\alpha(\Theta_\alpha,\mathcal{D}^b)$ and $\mathcal{L}_\beta(\Theta_\beta,\mathcal{D}^b)$:
\begin{equation}
\setlength{\abovedisplayskip} {0pt}
\setlength{\belowdisplayskip} {-5pt}
\mathcal{L}_\alpha(\Theta_\alpha,\mathcal{D}^b)={\mathbf{w}^b_\beta}^\top\bm{\ell}(\Theta_\alpha,\mathcal{D}^b),
\label{loss1}
\end{equation}
\begin{equation}
\setlength{\abovedisplayskip} {-2pt}
\setlength{\belowdisplayskip} {0pt}
\mathcal{L}_\beta(\Theta_\beta,\mathcal{D}^b)={\mathbf{w}^b_\alpha}^\top\bm{\ell}(\Theta_\beta,\mathcal{D}^b),
\label{loss2}
\end{equation}
where $\bm{\ell}(\Theta_\alpha,\mathcal{D}^b)$ and $\bm{\ell}(\Theta_\beta,\mathcal{D}^b)$ denote the loss values of the mini-batch $\mathcal{D}^b$ calculated by the network $\alpha$ and network $\beta$ respectively in the forward propagation phase. 

\begin{figure*}[!t]
	\setlength{\belowcaptionskip}{1 pt}
	\setlength{\abovecaptionskip}{2 pt}
	\centering
	\includegraphics[width=1.9\columnwidth]{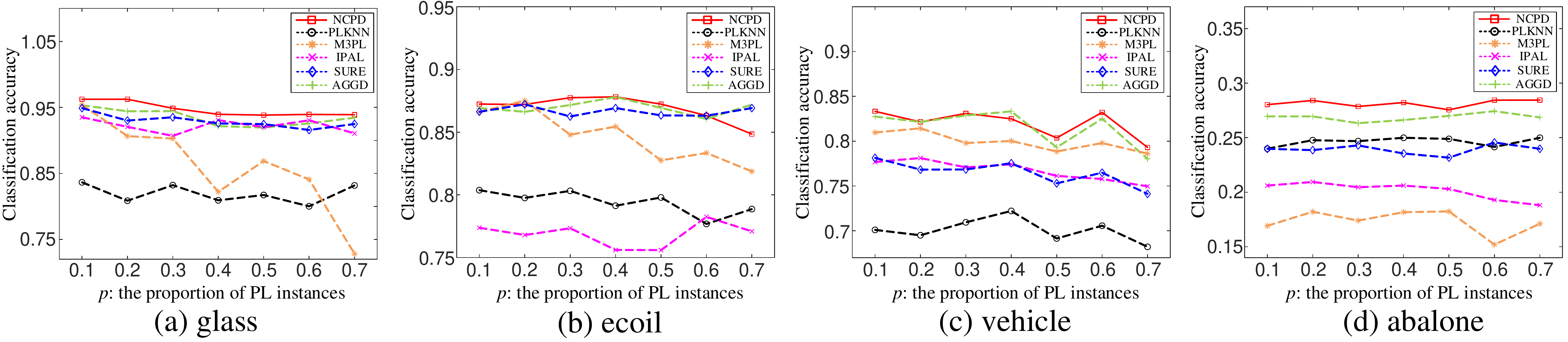}
	\caption{Classification accuracy of each algorithm on controlled UCI datasets with $p$ ranging from 0.1 to 0.7 ($r=1$).}
	\label{UCI_r1}
	\vskip -10pt 
\end{figure*}

\begin{figure*}[!t]
	\setlength{\belowcaptionskip}{1 pt}
	\setlength{\abovecaptionskip}{2 pt}
	\centering
	\includegraphics[width=1.9\columnwidth]{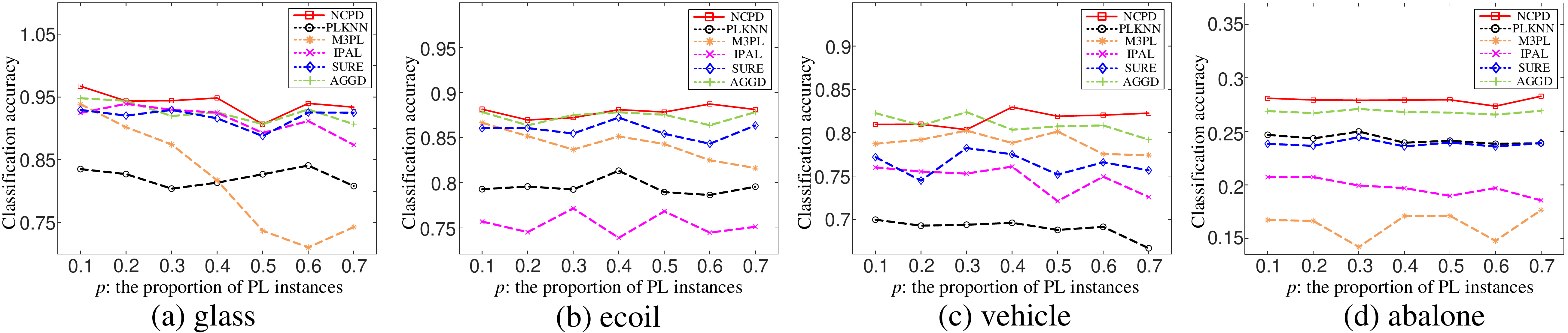}
	\caption{Classification accuracy of each algorithm on controlled UCI datasets with $p$ ranging from 0.1 to 0.7 ($r=2$).}
	\label{UCI_r2}
	\vskip -10pt
\end{figure*}

\begin{figure*}[!t]
	\setlength{\belowcaptionskip}{-3 pt}
	\setlength{\abovecaptionskip}{2 pt}
	\centering
	\includegraphics[width=1.9\columnwidth]{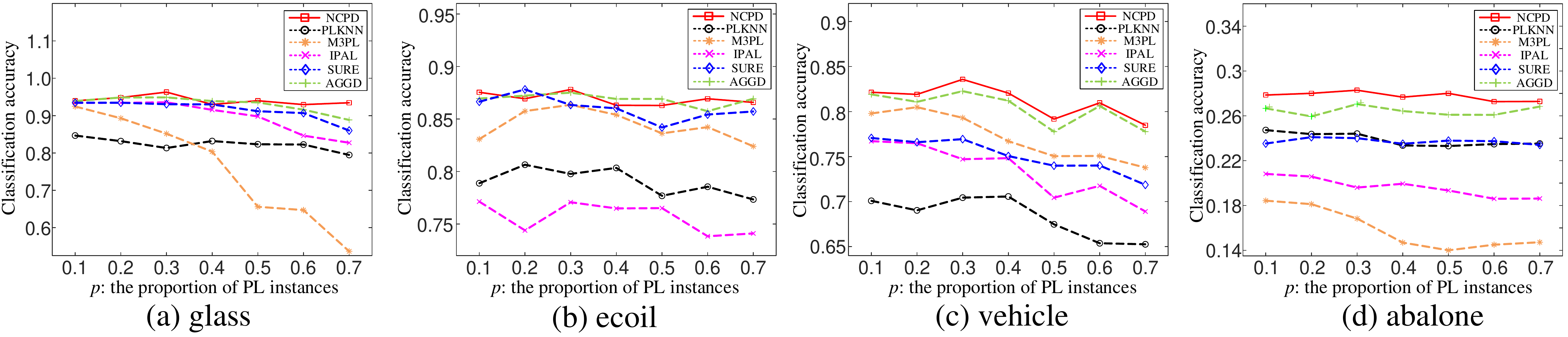}
	\caption{Classification accuracy of each algorithm on controlled UCI datasets with $p$ ranging from 0.1 to 0.7 ($r=3$).}
	\label{UCI_r3}
	\vskip -10pt
\end{figure*}

Eq.~(\ref{loss1}) and Eq.~(\ref{loss2}) indicate that each network exploits the data disambiguated by its peer network to train itself. As two networks have different ability and can disambiguate multi-birth groups at different levels, exchanging the confidence scores of instances is beneficial for both networks to reduce their respective disambiguation errors, and therefore the error accumulation problem inherited by the conventional single-trend training scheme can be effectively alleviated. Finally, we update the network parameters $\Theta_\alpha$ and $\Theta_\beta$ as follows:
\begin{equation}
\setlength{\abovedisplayskip} {0pt}
\setlength{\belowdisplayskip} {-9pt}
\Theta_\alpha:=\Theta_\alpha-\eta\nabla\mathcal{L}_\alpha(\Theta_\alpha,\mathcal{D}^b),
\end{equation}
\begin{equation}
\setlength{\abovedisplayskip} {0pt}
\setlength{\belowdisplayskip} {-10pt}
\Theta_\beta:=\Theta_\beta-\eta\nabla\mathcal{L}_\beta(\Theta_\beta,\mathcal{D}^b).
\end{equation}

\section{Experiments}
\label{experiments}
\subsection{Experimental Setup}
In this paper, we conduct comparative experiments to demonstrate the effectiveness of NCPD on two kinds of datasets, \emph{i.e.}, controlled UCI datasets and real-world partial label datasets. The compared state-of-the-art PLL algorithms include PLKNN~\cite{hullermeier2006}, M3PL~\cite{yu2017maximum}, IPAL~\cite{zhang2015solving}, SURE~\cite{feng2019partial}, and AGGD~\cite{wang2019adaptive}. 

For our NCPD approach, we employ the 3-layer perceptron as the backbone and meanwhile utilize Adam~\cite{kingma2014adam} to optimize the networks for all experiments. Besides, we employ the minibatch size of 128 and enable $T(t)$ to reach to one after 100 epochs, \emph{i.e.}, $t_r=100$, for all runnings. For baseline methods, they are implemented with parameters setup suggested in respective literatures. Specifically, the regularization parameter $C_{max}$ in M3PL is chosen from the set $\{0.01, 0.1, 1, 10, 100\}$ via cross-validation. In PLKNN, IPAL, and AGGD, the number of nearest numbers $k$ is chosen from the set $\{5,10,15,20\}$. Furthermore, we perform ten-fold cross-validation to record the mean prediction accuracies and standard deviations for all comparing algorithms on all the datasets adopted below.

\begin{table}[!t]
	\small
	\setlength{\belowcaptionskip}{-15pt}
	\setlength{\abovecaptionskip}{-3pt}
	\begin{center}
		\begin{tabular}{c|c|c|c|c}
			\hline
			\hline
			Datasets        & glass&  ecoil & vehicle & abalone\\
			\hline
			\# Instances       & 214 & 336      & 846  & 4,177 \\
			\hline
			\# Features        & 10   & 7      & 18    &7 \\
			\hline
			\# Labels          & 5   & 8      & 4    &29 \\
			\hline
			\hline
		\end{tabular}
		Configurations:\\
		(\uppercase\expandafter{\romannumeral1}) $r=1,p\in\{0.1,0.2,\cdots,0.7\}$\\
		(\uppercase\expandafter{\romannumeral2}) $r=2,p\in\{0.1,0.2,\cdots,0.7\}$\\
		(\uppercase\expandafter{\romannumeral3}) $r=3,p\in\{0.1,0.2,\cdots,0.7\}$
	\end{center}
	\caption{Characteristics and the parameter configurations of the controlled UCI datasets.}
	\label{UCI_datasets}
\end{table}

\begin{table*}[!t]
	\small
	\setlength{\belowcaptionskip}{-10pt}
	\setlength{\abovecaptionskip}{0pt}
	\setlength\tabcolsep{7.5pt}
	\begin{center}
		\begin{tabular}{c|c|c|c|c|c|c}
			\hline
			\hline
			{Datasets}     & NCPD                              &  PLKNN                            & M3PL                              & IPAL                              & SURE                             & AGGD\\
			\hline
			{Lost}         &\small{0.790 $\pm$ 0.055 }&\small{0.471 $\pm$ 0.032 $\bullet$}&\small{0.721 $\pm$ 0.037 $\bullet$}&\small{0.653 $\pm$ 0.022 $\bullet$}&\small{0.739 $\pm$ 0.036 $\bullet$}&\small{0.778 $\pm$ 0.040 }  \ \ \\
			\hline
			{BirdSong}     &\small{0.751 $\pm$ 0.018 }&\small{0.686 $\pm$ 0.015 $\bullet$}&\small{0.667 $\pm$ 0.042 $\bullet$}&\small{0.734 $\pm$ 0.013 $\bullet$}&\small{0.730 $\pm$ 0.015 $\bullet$}&\small{0.737 $\pm$ 0.018 } \ \  \\
			\hline
			{MSRCv2}       &\small{0.589 $\pm$ 0.046 }&\small{0.457 $\pm$ 0.049 $\bullet$}&\small{0.474 $\pm$ 0.038 $\bullet$}&\small{0.537 $\pm$ 0.045 $\bullet$}&\small{0.508 $\pm$ 0.043 $\bullet$}&\small{0.506 $\pm$ 0.041 $\bullet$}  \\
			\hline
			{Soccer Player}&\small{0.573 $\pm$ 0.013 }&\small{0.530 $\pm$ 0.016 $\bullet$}&\small{0.500 $\pm$ 0.007 $\bullet$}&\small{0.547 $\pm$ 0.016 $\bullet$}&\small{0.522 $\pm$ 0.013 $\bullet$}&\small{0.543 $\pm$ 0.016 $\bullet$}  \\
			\hline
			{Yahoo!News}   &\small{0.657 $\pm$ 0.013 }&\small{0.482 $\pm$ 0.011 $\bullet$}&\small{0.628 $\pm$ 0.013 $\bullet$}&\small{0.577 $\pm$ 0.010 $\bullet$}&\small{0.562 $\pm$ 0.011 $\bullet$}&\small{0.637 $\pm$ 0.008 $\bullet$}  \\
			\hline
			\hline
		\end{tabular}
		
	\end{center}
	\caption{Classification accuracy (mean $\pm$ std) of each algorithm on five real-world datasets. $\bullet / \circ$ indicates that NCPD is significantly superior / inferior to the comparing algorithm on the corresponding dataset (pairwise $t$-test with 0.05 significance level).}
	\label{results}
\end{table*}

\begin{table}[!t]
	\small
	\setlength{\belowcaptionskip}{-13pt}
	\setlength{\abovecaptionskip}{0pt}
	\setlength\tabcolsep{4.2pt}
	\begin{center}
		\begin{tabular}{c|c|c|c|c}
			\hline
			\hline
			Datasets        & \# Instances& \# Features & \# Labels & \# Avg. CLs  \\
			\hline
			Lost            & 1,122    & 108       & 16     & 2.23    \\
			\hline
			BirdSong        & 4,998    & 38        & 13     & 2.18    \\
			\hline
			MSRCv2          & 1,758    & 48        & 23     & 3.16    \\
			\hline
			Soccer Player   & 17,472   & 279       & 171    & 2.09   \\
			\hline
			Yahoo!News      & 22,991   & 163       & 219    & 1.91   \\
			\hline
			\hline
		\end{tabular}
		
	\end{center}
	\caption{Characteristics of adopted real-world partial label datasets.}
	\label{RealWorld_dataset}
\end{table}

\subsection{Experiments on Controlled UCI Datasets}
Following the widely-used controlling protocol in previous PLL works~\cite{cour2009learning,liu2012conditional,zhang2015solving,feng2019partial}, an artificial partial label dataset can be generated from an original UCI dataset with two controlling parameters $p$ and $r$. To be specific, $p$ controls the proportion of instances which are partially labeled (\emph{i.e.}, $|S_i|>1$), and $r$ controls the number of false positive labels in each candidate label set (\emph{i.e.}, $|S_i|=r+1$). The characteristics of these controlled UCI datasets as well as the parameter configurations are listed in Table~\ref{UCI_datasets}.

Figure~\ref{UCI_r1}, Figure~\ref{UCI_r2}, and Figure~\ref{UCI_r3} show the classification accuracy of each algorithm as $p$ ranges from 0.1 to 0.7 with the step size 0.1, when $r=1$, $r=2$, and $r=3$ (Configuration (\uppercase\expandafter{\romannumeral1}), (\uppercase\expandafter{\romannumeral2}), and (\uppercase\expandafter{\romannumeral3})), respectively. As illustrated in these figures, NCPD achieves superior performance against other comparing algorithms on these controlled UCI datasets. Specifically, NCPD achieves superior or at least comparable performance against PLKNN, M3PL, and IPAL in all experiments. As to SURE and AGGD, although their classification accuracies are slightly higher than NCPD in a few parameter configurations, they are inferior to NCPD in most cases.

\subsection{Experiments on Real-world Datasets}
Apart from the controlled UCI datasets, we also conduct experiments on five real-world partial label datasets which are collected from several application domains including \emph{Lost}~\cite{cour2009learning}, \emph{Soccer Player}~\cite{zeng2013learning}, and \emph{Yahoo!News}~\cite{guillaumin2010multiple} for automatic face naming, \emph{MSRCv2}~\cite{liu2012conditional} for object classification, and \emph{BirdSong}~\cite{briggs2012rank} for bird song classification. The characteristics of these real-world datasets are summarized in Table~\ref{RealWorld_dataset} where the average number of candidate labels of each dataset (\emph{i.e.}, \# Avg. CLs) is also reported.   

The average classification accuracies as well as the standard deviations of different approaches on these real-world datasets are shown in Table~\ref{results}. Pairwise $t$-test at 0.05 significance level is also conducted based on the results of ten-fold cross-validation. From Table~\ref{results}, we have three findings: 1) NCPD achieves the highest classification accuracies among all baselines on all adopted real-world datasets; 2) NCPD significantly outperforms PLKNN, M3PL, IPAL, and SURE on all these datasets; 3) NCPD is never statistically inferior to any comparing algorithms in all cases. These findings convincingly substantiate the superiority of our NCPD approach to other comparators.

\begin{figure}[!t]
	\setlength{\belowcaptionskip}{-3 pt}
	\setlength{\abovecaptionskip}{1 pt}
	\centering
	\includegraphics[width=0.63\columnwidth]{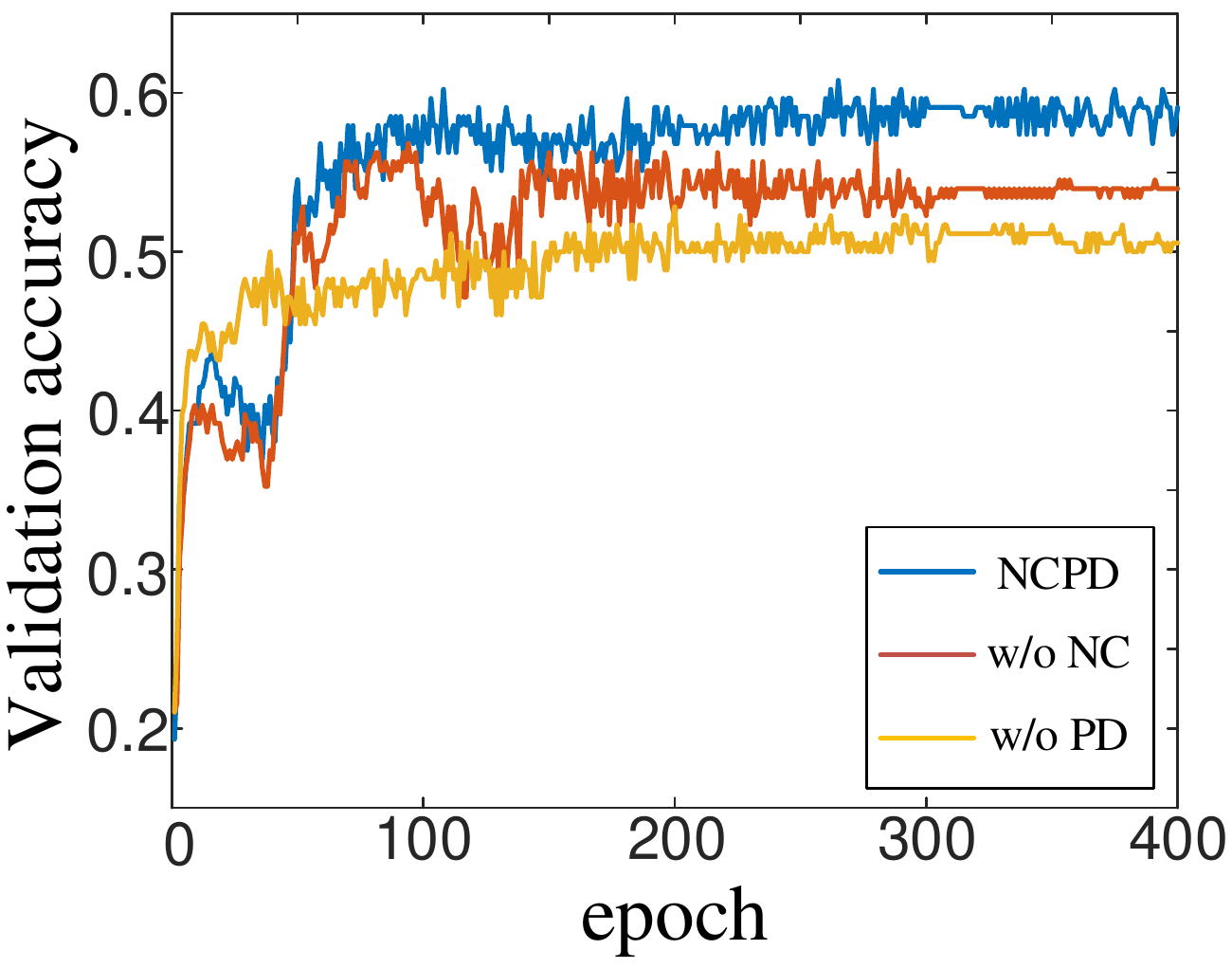}
	\caption{Validation accuracy with different settings on \emph{MSRCv2} dataset. The blue curve denotes the accuracy of the integrated NCPD approach (legend by ``NCPD''). The red curve and the yellow curve indicate the accuracy of NCPD that removes the network cooperation mechanism (denoted by ``w/o NC'') and the progressive disambiguation strategy (denoted by ``w/o PD''), respectively.}
	\label{acc}
	\vskip -10pt
\end{figure}

\subsection{Ablation Study}
The superiority of the proposed NCPD approach has been verified by thorough experimental results presented above. In this section, we conduct ablation study on \emph{MSRCv2} dataset to further demonstrate the effectiveness of the two crucial techniques employed by NPCD, \emph{i.e.}, the progressive disambiguation strategy and the network cooperation mechanism. 

Specifically, to demonstrate the effectiveness of the progressive disambiguation strategy, we discard this strategy and merely train two networks with network cooperation mechanism, \emph{i.e.}, all multi-birth groups are disambiguated according to Eq.~(\ref{map1}) in every epoch regardless their disambiguation difficulty. To confirm the effectiveness of the network cooperation mechanism, we barely train one network equipped with the progressive disambiguation strategy (see Section~\ref{3.2}). Figure~\ref{acc} shows the results, from which we can observe that the integrated NCPD approach generates the highest accuracy than other two settings (\emph{i.e.}, ``w/o NC'' and ``w/o PD''). In contrast, the accuracy will decrease when either the progressive disambiguation strategy or the network cooperation mechanism is removed, therefore the effectiveness and indispensability of these two crucial techniques are validated.

\section{Conclusion}
\label{conclusion}
In this paper, we propose a novel approach for PLL which is dubbed as ``NCPD''. By employing the progressive disambiguation strategy, our approach is able to exploit the disambiguation difficulty of the instances and then disambiguate them in a progressive manner, which is beneficial for the steady improvement of model capability and thereby the adverse impacts brought by false positive labels can be effectively reduced. Furthermore, the network cooperation mechanism greatly facilitates the salutary mutual learning process between two networks, and therefore can effectively alleviate the error accumulation problem inherited by the existing single-trend training framework. Thorough experimental results on various datasets demonstrate the effectiveness of the proposed NCPD approach. Considering that how to determine the disambiguation difficulty of the instances plays a vital role in our algorithm, we will devise a more advanced methodology to judge the disambiguation difficulty of these partially labeled instances in the future.

\newpage

\bibliographystyle{named}
\bibliography{ijcai20}

\end{document}